\documentclass[sigconf]{acmart}

\usepackage{cleveref}
\usepackage{subcaption}
\usepackage{graphicx} 
\usepackage{makecell}

\AtBeginDocument{%
  \providecommand\BibTeX{{%
    \normalfont B\kern-0.5em{\scshape i\kern-0.25em b}\kern-0.8em\TeX}}}

\setcopyright{acmlicensed}
\copyrightyear{2022}
\acmYear{2022}
\acmDOI{10.1145/3491140.3528329}

\acmConference[L@S '22]{Proceedings of the Ninth ACM Conference on Learning @ Scale}{June 1--3, 2022}{New York City, NY, USA}
\acmBooktitle{Proceedings of the Ninth ACM Conference on Learning @ Scale (L@S '22), June 1--3, 2022, New York City, NY, USA}
\acmPrice{15.00}
\acmISBN{978-1-4503-9158-0/22/06}
\begin{document}
\fancyhead{}
\title{Using Chatbots to Teach Languages}


\author{Yu Li}
\affiliation{%
  \institution{Columbia University}
  \city{New York City}
  \country{United States}
}
\email{yl5016@columbia.edu}

\author{Chun-Yen Chen}
\affiliation{%
  \city{New York City}
  \country{United States}
}
\email{itsarbit@gmail.com}

\author{Dian Yu}
\affiliation{%
  \institution{University of California, Davis}
  \city{Davis}
  \country{United States}
}
\email{diayu@ucdavis.edu}

\author{Sam Davidson}
\affiliation{%
  \institution{University of California, Davis}
  \city{Davis}
  \country{United States}\\
}
\email{ssdavidson@ucdavis.edu}

\author{Ryan Hou}
\affiliation{%
  \institution{Columbia University}
  \city{New York City}
  \country{United States}
}
\email{rh2920@columbia.edu}

\author{Xun Yuan}
\affiliation{%
  \institution{Zhejiang University}
  \city{Hangzhou}
  \country{China}\\
  yuanxun@zju.edu.cn
}

\author{Yinghua Tan}
\affiliation{%
  \city{New York City}
  \country{United States}
}
\email{luckyfafa214@gmail.com}

\author{Derek Pham}
\affiliation{%
  \institution{Columbia University}
  \city{New York City}
  \country{United States}
}
\email{dp3081@columbia.edu}

\author{Zhou Yu}
\affiliation{%
  \institution{Columbia University}
  \city{New York City}
  \country{United States}
}
\email{zy2461@columbia.edu}

\begin{abstract}
  This paper reports on progress towards building an online language learning tool to provide learners with conversational experience by using dialog systems as conversation practice partners. Our system can adapt to users’ language proficiency on the fly. We also provide automatic grammar error feedback to help users learn from their mistakes. According to our first adopters, our system is entertaining and useful. Furthermore, we will provide the learning technology community a large-scale conversation dataset on language learning and grammar correction. Our next step is to make our system more adaptive to user profile information by using reinforcement learning algorithms.
\end{abstract}



\maketitle


\begin{figure*}
    \begin{subfigure}[t]{.38\linewidth}
    \hbox{\hspace{1em}
    \includegraphics[width=\linewidth]{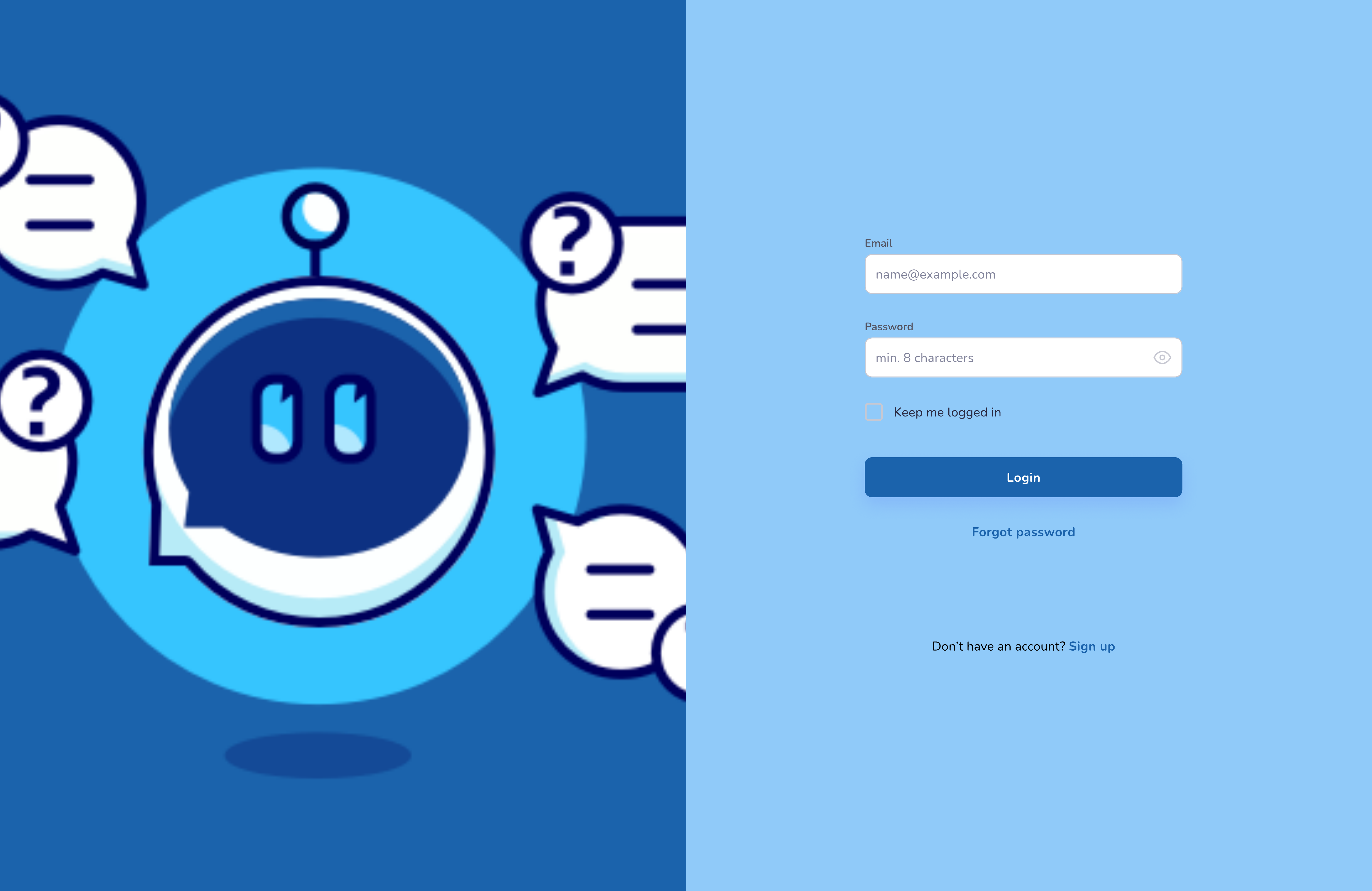}}
    \caption{login page}
    \end{subfigure}
\hfill
    \begin{subfigure}[t]{.38\linewidth}
    \hspace{-1em}
    \includegraphics[width=\linewidth]{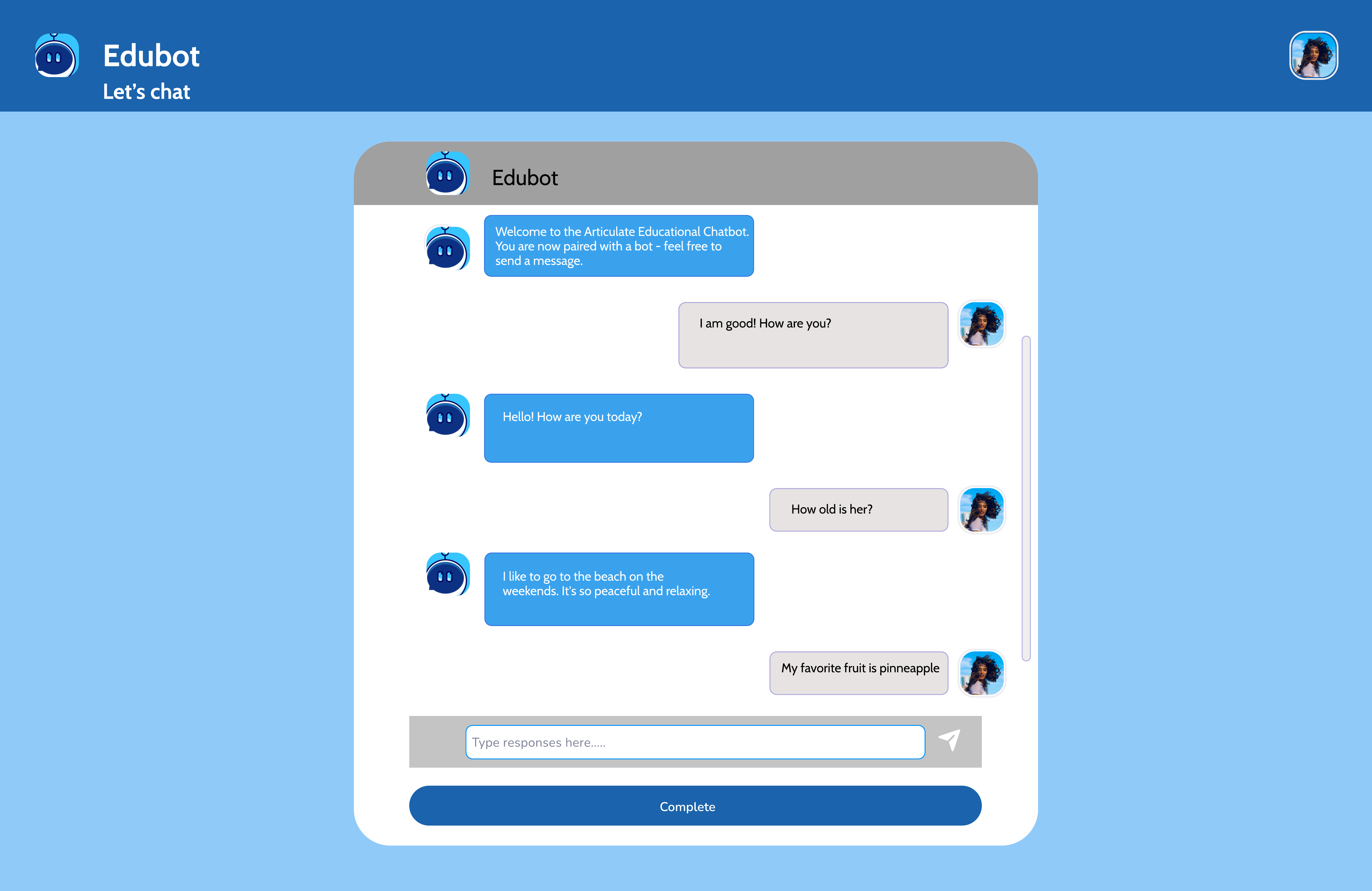}
    \caption{conversation practice}
    \end{subfigure}
\hfill
    \begin{subfigure}[t]{.38\linewidth}
    \hbox{\hspace{1em}
    \includegraphics[width=\linewidth]{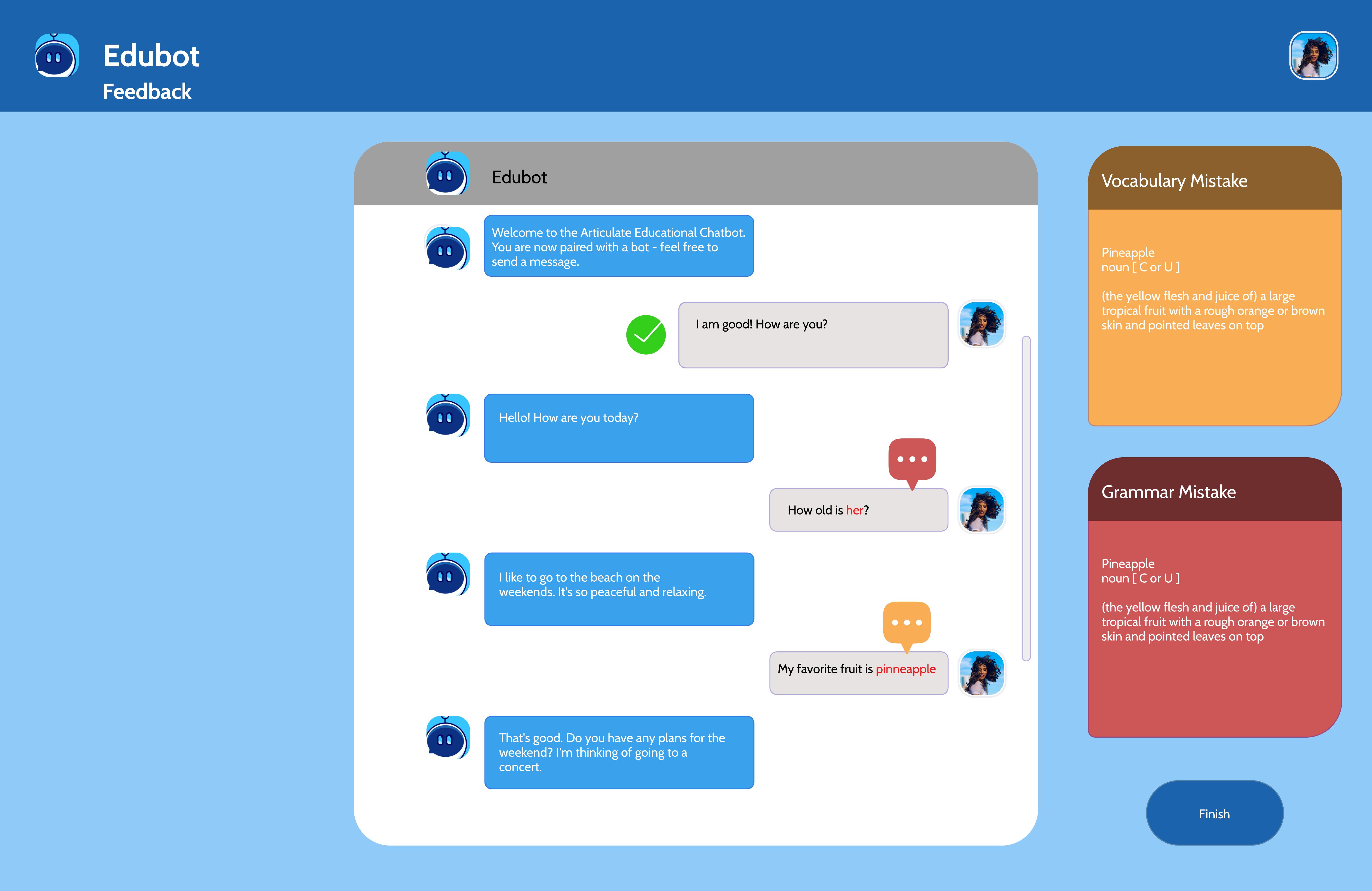}}
    \caption{grammar error feedback}
    \end{subfigure}
\hfill
    \begin{subfigure}[t]{.38\linewidth}
    \hspace{-1em}
    \includegraphics[width=\linewidth]{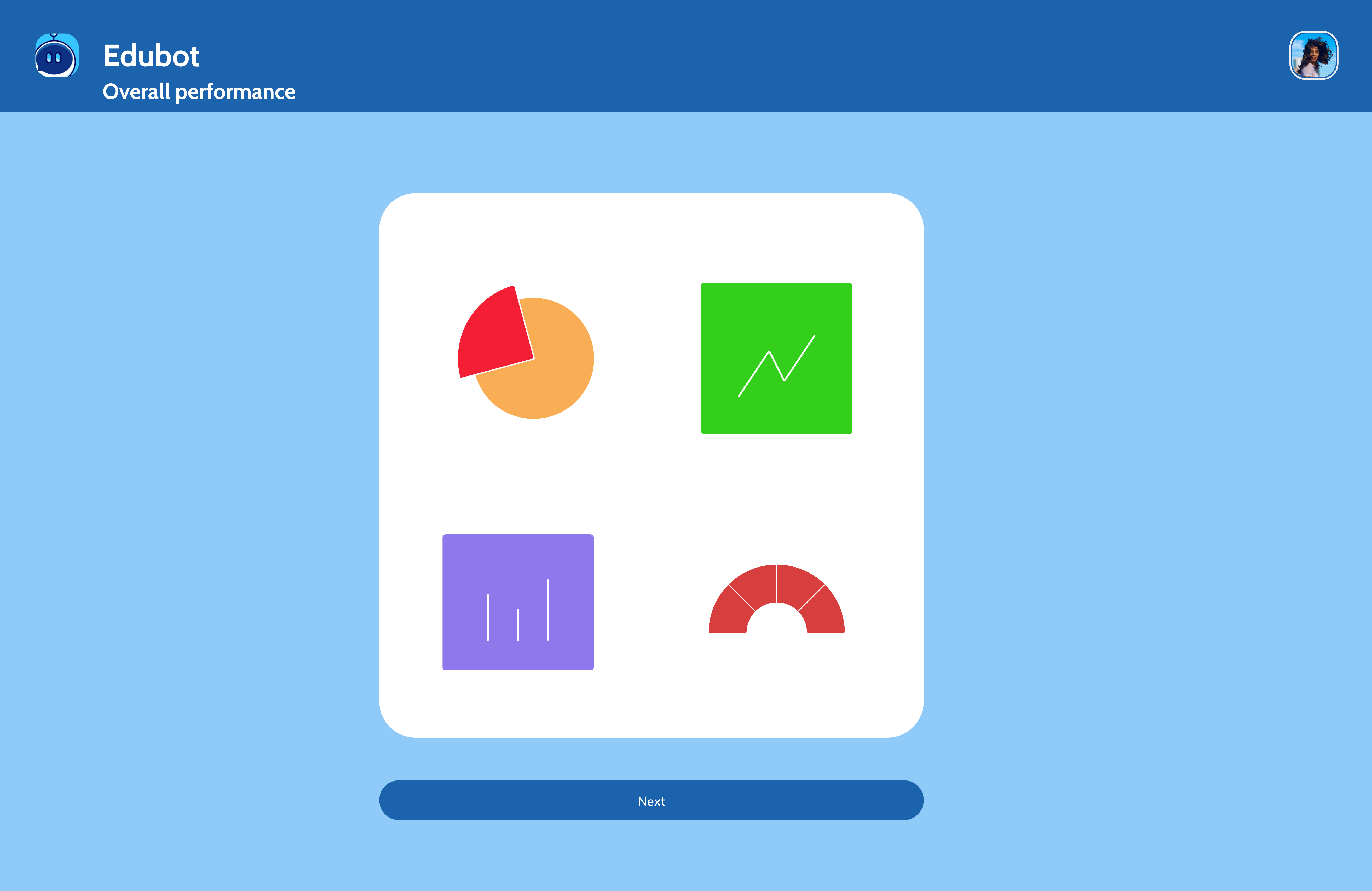}
    \caption{overall performance}
    \end{subfigure}
\caption{Screenshots of our tool. It includes login pages, conversation practice, grammar error feedback, and long-term user overall performance.}
\label{fig:data_coll}
\vspace{-5mm}
\end{figure*}

\section{Introduction}
It is estimated that over one billion people are currently learning English worldwide. Nowadays, language learning platforms such as Duolingo\footnote{https://www.duolingo.com/} have become popular worldwide . Most of these applications focus on correcting learners' vocabulary, grammar, and writing, but do not emphasize one of the critical aspects of acquiring a new language: conversational skills that make you sound naturally. Traditionally, such conversational skills are acquired through in-class communication with a qualified instructor, who may not necessarily be a native speaker, or by conversing with a native speaker of the target language. Unfortunately, many learners do not have access to a qualified instructor or native speaker to practice their conversational skills in their target language. Even when they are in an environment with native speakers, many learners experience communication anxiety when conversing in a new language, often due to their low-level speaking skills or cultural differences \cite{Macintyre2007WillingnessTC}. Foreign language learning theories postulate that ample practice is required for fluent speaking \cite{dekeyser_dekeyser_2007}. However, limited teaching resources are available to meet this need, and current language learning technologies, which seek to democratize language learning, cannot fill the gap. We seek to fill this gap in current language learning technologies using an AI-powered educational dialog system for language learning which has the ability to adapt to individual learners' language proficiency.

Educational dialog systems are not as popular as personal assistants due to limited investment and the underlying difficulty of developing effective conversational dialog systems. As a result of these challenges, most educational dialog systems react to each learner with a similar conversation flow without considering learner proficiency or preferences. Thus, interactions tend to be tedious and repetitive, making it ineffective in engaging learners and providing a personalized learning experience. However, long-term and frequent use is essential for effective language learning. Therefore, learner engagement and enjoyment are even more of a concern for our proposed system than for conversational dialog systems in general. We propose to design end-to-end trainable, user-adaptive dialog systems to teach second-language communication skills with both typing and spoken settings. Our systems provide highly engaging, personalized, and effective language education at scale with a low cost, making learning accessible to the general public.

Our tool provides language learners with a conversational experience using dialog systems as conversation practice partners. The chatbot system follows a sound social chatbot \cite{liang2020gunrock}. It achieved state-of-the-art results in interacting with massive numbers of actual users in having engaging and interesting conversations. Moreover, we provide automatic grammar feedback after each conversation to help learners learn from their mistakes. Figure \ref{fig:data_coll} shows some snapshots of our tool. Specifically, we develop a model for automatic grammar correction and show learners an annotated list of their types of grammatical mistakes, elicit self-correction, and provide a list of directions in spoken narrations to help learners improve their English language proficiency.

In addition, our conversations follow a rigorous curriculum that can help learners improve their language proficiency step-by-step and be combined with in-classroom education. This curriculum will allow the conversational system to create more opportunities for learners to practice errors they frequently make. For example, if a learner makes past tense errors frequently, based on the automatic grammar correction model's output, we will adjust the dialog system to create scenarios that allow learners to practice past tense more. An example system utterance could be: "Could you tell me more about the Taylor Swift concert you watched online? What was Taylor wearing?" Finally, we will develop a visualization interface to track learners' improvement over different learning criteria, including grammar, vocabulary, and sentence structures.

In the planned work, we expect to provide language learning community with a new open-source framework for dialog policy training. We will also provide the research community with a dataset of conversations with annotations of grammar error types and corrected grammar errors, along with a trained grammar correction model and a grammar type detector. Most importantly, our system's output, a second language communication training system, would bring new opportunities for second language learners to acquire training through interacting with our system online. Such a system is a valuable environment for learners to practice the language with a ``native speaker'' without the same social anxieties.

\section{Related Work}
Language teachers have brought bots into the foreign language learning classroom as a permanent tool for language practice \citet{fryer2006bots}.
In education research, there are dialog systems for physics tutoring \cite{Mathews2003}, language learning \cite{bygate2013researching, fryer2019chatbot}, and even communication training \cite{Yu2019}. However, no system has been evaluated or deployed on popular device platforms that have the potential of providing access to real learners. In this work, we propose to design, implement, and deploy adaptive dialog systems to a specific educational task, second language conversational communication training and to deploy it on popular platforms to gain wider access. 

Research in second language learning has explored numerous methods of providing feedback to students to inform them of errors in writing or speech. The most basic form of feedback, often referred to as ``implicit'' feedback, consists of using the correct form of an error made by an interlocutor; this is the type of feedback that children largely rely on when learning their first language. Second language acquisition research, on the other hand, suggests that second language learners generally benefit more from explicit feedback than from implicit feedback, as shown in several syntheses and meta-analyses for various aspects of language learning: pragmatics/appropriate politeness strategies \cite{jeon2006effects, taguchi_2015}, grammar \cite{Norris2002Effectiveness, Li2010Effectiveness}, and vocabulary \cite{Dilans2010Corrective} (not a meta-analysis or a synthesis). However, explicit feedback can take many forms that focus on various aspects of linguistic competence, such as grammar, lexicon, or pragmatics. The best type of feedback, or combination of feedback types, to provide to students remains an open question. For example, \citet{Bardovi2006Congruence} argued that it is pragmatics, rather than grammatical competence, which distinguishes non-native learners from native speakers. In this study, we test the effects of two types of explicit feedback: 1) meta-pragmatic suggestions provided to learners during the conversation, and 2) explicit grammatical feedback provided after the conversation ends. Our goal is to determine which type of feedback will result in more native-like conversational competence.

There are different levels of adaptation the system could do with respect to individual users. \citet{janarthanam-lemon-2014-adaptive} proposed to adapt dialog systems according to the user’s domain knowledge in a help system to make systems more effective. \citet{Thompson2004Personalized} utilized user preferences for recommendation dialog systems. \citet{Yu2016UserES, TimpeLaughlinDombi2020} discussed how to build systems respecting users from different cultures. In this study, we would utilize the users’ profile information, specifically the language skills and learning experience of the users. We plan to make further adaptations on different levels in our future work.

\begin{table*}[htb!]
    \centering
    \small
    \begin{tabular}{p{8.5cm}p{8.5cm}}
    \textbf{Rule-based Adaptive Policy} & \textbf{RL-based Adaptive Policy}\\
    \midrule
    \midrule
    \textbf{System}: What would you do if you have a disagreement with a co-worker? Would you settle the matter directly with your co-worker or would you report the incident to your manager?\newline
    \textbf{User}: I’d talk to co-worker, so he knows what he is doing and he doesn’t get into real trouble. Well if he do something really really bad, then maybe some other person should tell the manager.\newline
    \textbf{System}: What would you think the manager would do next?\newline
    \textbf{User}: Ahh, cut his pay a little bit. I don’t really know actually. \textcolor{orange}{[negative sentiment \& uncertainty based on low voice and disfluency.]}\newline
    \textbf{System}: It is more effective for the manager to call for a meeting with the three of you to talk openly about the issue. \textcolor{red}{[provide one correct answer]} How would you deal with an unreasonable demand by a customer? \textcolor{red}{[next scenario]}

    & \textbf{System}: ...\newline
    \textbf{User}: ...\newline
    \textbf{System}: What would you think the manager would do next?\newline
    \textbf{User}: Ahh, cut his pay a little bit. I don’t really know actually. \textcolor{orange}{[negative sentiment \& uncertainty based on low voice and disfluency.]}\newline
    \textbf{System}: Not a bad idea, cutting one’s pay is one way to deal with the situation. \textcolor{red}{[flexible encouragement strategy to deal with uncertainty and low language proficiency]}. Would you think it would be helpful if the manager called for a meeting with the three of you to talk openly about the issue? \textcolor{red}{[provide one correct answer  with a suggestive voice]}\newline
    \textbf{User}:  Yes. That’s a good idea. \textcolor{blue}{Displayed hints: Talk more about how to actively mediate communication.}\newline
    \textbf{System}: I’m glad that you agree with me.  How would you deal with an unreasonable demand by a customer? \textcolor{red}{[next scenario]}\\
    \midrule
    \multicolumn{2}{l}{\textcolor{green}{Explicit Feedback:}}\\
    \textbf{Grammar errors}: & \textbf{Content suggestions}:\\
    Missing determiner: talk to \textbf{the} co-worker. & It would be great to talk about the conflict openly.\\
    Subject-verb disagreement: Well, if he \textbf{did} something & \\
    \\
    \end{tabular}
    \caption{An example dialog of rule-based uncertainty adaptive systems and an expected RL-based adaptive policy with in-task pragmatic hints and after-task explicit feedback on both grammar and content. }
    \label{tab:example}
    \vspace{-9mm}
\end{table*}

\section{Approach}
In this section, we will describe our dialog system framework (\ref{app:dialog_system}) and automatic feedback system (\ref{app:feedback}).

\subsection{Dialog System}
\label{app:dialog_system}
Our chatbot follows state-of-the-art social chatbot \cite{liang2020gunrock}. To improve the system for educational purposes, we are in the process of designing a dialog policy that incorporates user profile information.

Incorporating user information about their language proficiency in dialog model training will help the system to adapt to individual users and make users feel the conversation is personalized. We expect users will engage with the customized system more, improve their confidence, and have an overall better learning gain. We will first design the dialog policy so that, during training, it is conditioned on the user’s profile information. This specific study uses the following information: the number of years spent learning the language and their first language. Our targeted application is English. According to a previous study \cite{kellerman_1979}, when learners’ first language is similar to the targeted language, they usually exhibit better English communication skills in general. Algorithmically, we will add these two user attributes in the policy gradient method employed in training the reinforcement learning policy. The states for reinforcement learning are the dialog states pre-designed by experts. A dialog state is in the form of: ``Whether has previous experience on disagreement with a co-worker, actions taken in handling disagreement with a co-worker''. The dialog system actions are also a list of finite actions predefined by the expert, such as ``Request previous experience on disagreement with a co-worker, request more details''. We use ARDM \cite{Wu2021Alternating} as the base for the language generation model for the dialog system. Before training the model with a user simulator, we first pre-train it with existing social chat corpora, blending dialog skill tasks to achieve a fluent dialog language model. The reason that we use a discrete number of states predefined is due to quality control. Since this is an integration proposal, we aim to ensure the dialog system is well functioned instead of proposing an innovative but risky new dialog system model. Although our system requires an expert’s effort to design the state and action space, it is still end-to-end trainable, similar to \citet{shi-yu-2018-sentiment}. This work will integrate profile information in reinforcement learning by including them in state design. We will also include profile information prior to the reinforcement learning model to reduce the optimization steps. We plan to explore reducing the expert involvement in dialog systems design in our future work.

To build our reinforcement learning-based (RL-based) adaptive policy, we will first collect data consisting of human resource experts interviewing users as potential job candidates. We will constrain the interview’s flow by asking them to focus on a semi-limited flow. We show an example of the original conversation task we developed in Table \ref{tab:example}, in which the system utilizes a rule-based adaptive policy on the left, while an expected RL-based adaptive conversation example is shown on the right. The rule-based system has only a pre-designed template. It cannot detect whether the user’s answer is correct and only responds with a pre-written template for encouragement if the user has uncertainty. For instance, at the last turn, the user is hesitating, the sentiment and uncertainty detector has detected that the user is disengaged, and the system then switches to a branch that gives encouragement and then immediately moves to the following scenario. However, the system only uses a strategy to handle the user disengagement but does not deal with the fact that we need to give the user some good sample responses to learn from for such a training system. However, the RL-based system can utilize a reinforcement learning trained dialog policy to select from a list of templates pre-rewritten by human HR experts. Both sentiment and uncertainty will be the intermediate reward of the reinforcement learning method. The RL-based policy would learn the optimal path that utilizes all the pre-designed good interviewer behaviors to help users learn efficiently. In contrast, the rule-based system can only add limited extra branches to handle various user responses due to the rigidity of the rules.

\subsection{Automatic Feedback Design}
\label{app:feedback}
We have developed a model for automatic grammar correction and have been developing other automatic language proficiency metrics, such as syntactic complexity and diversity, vocabulary size, and overall language fluency. Existing models have already been developed for written essays but not for conversational settings. In conversations, models must treat fragments, ellipsis, and repetition (e.g., very very good) differently from written texts. Thus, we have to retrain or finetune previous models on conversational data; we compile the necessary data in our preliminary conversation collections since no conversational grammatical error correction corpus is currently available. The development of this error correction model will allow us to show users their conversations with marked errors, elicit self-correction, and give them a list of things they could work on in natural language narrations to help users improve their English language proficiency. In addition, we will also provide a 1-5 Likert score to users based on their overall performance to help them track their progress. This score comes from a supervised learning classification model on a set of conversations with performance ratings. We gather the needed conversation data using Amazon Mechanical Turkers. Specifically, we hire crowd workers who are non-native speakers to interact with our systems, and we will ask language learning experts from ETS \footnote{https://www.ets.org/} to provide overall proficiency scores on these conversations.

\section{Results}
We collected 120 conversations using our tool on Amazon Mechanical Turk platform from 100 users who chatted with it. We asked users to rank the usefulness of our chatbot on a 1-5 Likert scale, with five meaning instrumental, and our chatbot achieved an average score of 4.3. We also asked users if they would recommend the bot to a friend or family; we found 87 \% replied "yes." 84\% of people feel they are more confident after the interaction. We elicit feedback from users through open questions in an end-of-task survey. Some users give encouraging feedback such as "It is good to have a not-judging conversational partner," and "I find the grammar correction useful." There are also users giving suggestions to improve the system further, such as "The bot does not listen to me." To integrate this feedback, we will make the system more adaptive to an individual's responses by using conditional neural network generation models, where the condition is active listening strategies.

\section{Discussion}
Language generation safety is a critical issue in end-to-end generation-based conversational systems. Because the generation process is a stochastic decoding process, the generated sentences may contain unpredictable content that may violate common sense or even contain hate speech and profanity. We use profanity and hate speech detectors trained using social conversational data to filter out potential invalid responses to avoid such risk.

Another issue with the end-to-end dialog generation model is how to keep conversations consistent over time and avoid logical conflicts. We will apply a two-pass framework. In the first pass, we use a pre-trained reading comprehension model to detect contradiction and repetition. Then we use the detected errors as filters and only output valid candidates \cite{2021shiaccented}.

\section{Conclusion and Future Work}
This paper presents an online language learning tool to provide learners with conversational experience using dialog systems as conversation practice partners. Our system supports both typing, and spoken interactions in a mixed modularized and end-to-end dialog system. Our system also provides grammar error feedback to help users learn from their mistakes. It brings new opportunities for second language learners to acquire training through active interaction at a low cost. According to our first adopters, over 87\% of users would recommend the bot to a friend or family.

We will develop a reinforcement learning-based adaptive policy in our system in the future. We will offer more advanced features, such as a personalized curriculum and a more extensive set of conversational topics for users. We also plan to collaborate with other online learning platforms, such as Udemy \footnote{https://www.udemy.com/}, to help them design specific conversational agents targeted at specific teaching materials.

\section{Acknowledgements}
Thanks to Dr. Luke Fryer and Max Chen for their helpful feedback and discussions.
\bibliographystyle{ACM-Reference-Format}
\bibliography{custom}

\appendix
\end{document}